\pdfoutput=1
\documentclass[11pt,a4paper]{article}
\usepackage[hyperref]{naaclhlt2019}
\usepackage{times}
\usepackage{latexsym}
\usepackage{url}

\usepackage{amssymb}
\usepackage{amsmath}
\usepackage{latexsym}
\usepackage{url}
\usepackage{mathtools}
\usepackage{multirow}
\usepackage{xspace}
\usepackage{comment}
\usepackage{flowchart}
\usetikzlibrary{arrows}
\usepackage{booktabs}
\usepackage{subcaption}
\usepackage[algo2e,vlined,ruled]{algorithm2e}
\usepackage{tikz}
\usepackage{pgfplots}

\DeclareMathAlphabet{\pazocal}{OMS}{zplm}{m}{n}
\DeclareMathAlphabet{\pazocal}{OMS}{zplm}{m}{n}

\aclfinalcopy 

\newcommand{\red}[1]{\textcolor{red}{#1}}

\newcommand{\ie}{{\em i.e.,}\xspace}
\newcommand{\eg}{{\em e.g.,}\xspace}
\newcommand{\aka}{\emph{a.k.a.}\xspace}

\newcommand{\resp}{{res.}\xspace}

\newcommand{\Xs}{\pazocal{X}}
\newcommand{\Ys}{\pazocal{Y}}
\newcommand{\Zs}{\pazocal{Z}}

\newcommand{\Ni}{({\em i})~}
\newcommand{\Nii}{({\em ii})~}

\newcommand{\Na}{({\em a})~}
\newcommand{\Nb}{({\em b})~}
\newcommand{\Nc}{({\em c})~}
\newcommand{\Nd}{({\em d})~}
\newcommand{\Ne}{({\em e})~}
\newcommand{\Nf}{({\em f})~}

\title{Revisiting Adversarial Autoencoder for Unsupervised Word Translation  with Cycle Consistency and Improved Training}

\author{Tasnim Mohiuddin$^\star$ \and Shafiq Joty$^\star$$^\dagger$ \\
$^\star$Nanyang Technological University, Singapore \\
$^\dagger$Salesforce Research Asia, Singapore \\
{\tt \{mohi0004,srjoty\}@ntu.edu.sg}
\\}

\date{}

\begin{document}
\maketitle

\begin{abstract}

Adversarial training has shown impressive success in learning bilingual dictionary without any parallel data by mapping monolingual embeddings to a shared space. However, recent work has shown superior performance for non-adversarial methods in more challenging language pairs. In this work,  we revisit adversarial autoencoder for unsupervised word translation and propose two novel extensions to it that yield more stable training and improved results. Our method includes regularization terms to enforce cycle consistency and input reconstruction, and puts the target encoders as an adversary against the corresponding discriminator. Extensive experimentations with European, non-European and low-resource languages show that our method is more robust and achieves better performance than recently proposed adversarial and non-adversarial approaches. 


\end{abstract}

\section{Introduction} 
\label{sec:intro}

Learning cross-lingual word embeddings has been shown to be an effective way to transfer knowledge from one language to another for many key linguistic tasks including machine translation, named entity recognition, part-of-speech tagging, and parsing \cite{ruder2017survey}. While earlier efforts solved the associated word alignment problem using large parallel corpora \cite{Luong15-bivec}, broader applicability demands methods to relax this requirement since acquiring a large corpus of parallel data is not feasible in most scenarios. Recent methods instead use embeddings learned from monolingual data, and learn a linear mapping from one language to another with the underlying assumption that two embedding spaces exhibit similar geometric structures (\ie\ approximately \emph{isomorphic}). This allows the model to learn effective cross-lingual representations without expensive supervision \cite{artetxe2017acl}.

Given monolingual word embeddings of two languages, \newcite{Mikolov13} show that a linear mapping can be learned from a seed dictionary of 5000 word pairs by minimizing the sum of squared Euclidean distances between the mapped vectors and the target vectors. Subsequent works \cite{XingWLL15,artetxe2016emnlp,artetxe2017acl,SmithICLR17} propose to improve the model by normalizing the embeddings, imposing an orthogonality constraint on the mapper, and modifying the objective function. While these methods assume some supervision in the form of a seed dictionary, recently fully unsupervised methods have shown competitive results. \newcite{Zhang17,Zhang-17-emnlp} first reported encouraging results with \emph{adversarial training}. \newcite{conneau2018word} improved this approach with post-mapping refinements, showing impressive results for several language pairs. Their learned mapping was then successfully used to train a fully unsupervised neural machine translation system \cite{lample2017unsupervised,lample2018phrase}. 

Although successful, adversarial training has been criticized for not being stable and failing to converge, inspiring  researchers to propose \emph{non-adversarial} methods more recently \cite{Xu2018,Hoshen-18,david2018gromov,Artetxe-2018-acl}. In particular, \newcite{Artetxe-2018-acl}  show that the adversarial methods of \newcite{conneau2018word} and \newcite{Zhang17,Zhang-17-emnlp} fail for many language pairs.     

In this paper, we revisit adversarial training and propose a number of key improvements that yield more robust training and improved mappings. Our main idea is to learn the cross-lingual mapping in a projected latent space and add more constraints to guide the unsupervised mapping in this space. We accomplish this by proposing a novel \emph{adversarial autoencoder} framework \cite{Makhzani2015AdversarialA}, where adversarial mapping is done at the (latent) code space as opposed to the original embedding space (Figure \ref{fig:proposed-model}). This gives the model the flexibility to automatically induce the required geometric structures in its latent code space that could potentially yield better mappings. \newcite{Anders-18} recently find that the \emph{isomorphic} assumption made by most existing methods does not hold in general even for two closely related languages like English and German. In their words \emph{``approaches based on this assumption have important limitations''}. By mapping the latent vectors through adversarial training, our approach therefore departs from the isomorphic assumption. 

In our adversarial training, not only the mapper but also the target encoder is trained to fool the discriminator. This forces the discriminator to improve its discrimination skills, which in turn pushes the mapper to generate indistinguishable translation. To guide the mapping, we include two additional constraints. Our first constraint enforces \emph{cycle consistency} so that code vectors after being translated from one language to another, and then translated back to their source space remain close to the original vectors. The second constraint ensures reconstruction of the original input word embeddings from the back-translated codes. This grounding step forces the model to retain word semantics during the mapping process. 

We conduct a series of experiments with six different language pairs (in both directions) comprising European, non-European, and low-resource languages from two different datasets. Our results show that our model is more robust and yields significant gains over \newcite{conneau2018word} for all translation tasks in all evaluation measures. Our method also gives better initial mapping compared to other existing methods \cite{Artetxe-2018-acl}. We also perform an extensive ablation study to understand the contribution of different components of our model. The study reveals that cycle consistency contributes the most, while adversarial training of the target encoder and post-cycle reconstruction also have significant effect. 
We have released our source code at \url{https://ntunlpsg.github.io/project/unsup-word-translation/}

The remainder of this paper is organized as follows. After discussing related work in Section~\ref{sec:rel-work}, we present our unsupervised word translation approach with adversarial autoencoder in Section~\ref{sec:model}. We describe our experimental setup in Section~\ref{sec:setting}, and present our results with in-depth analysis in  Section~\ref{sec:result}. Finally, we summarize our findings with possible future directions in  Section~\ref{sec:conclusion}. 




\section{Related Work}
\label{sec:rel-work}
In recent years a number of methods have been proposed to learn bilingual dictionary from monolingual word embeddings.\footnote{see \cite{ruder2017survey} for a nice survey} Many of these methods use an initial seed dictionary. \newcite{Mikolov13} show that a linear transformation can be learned from a seed dictionary of $5000$ pairs by minimizing the squared Euclidean distance. In their view, the key reason behind the good performance of their model is the similarity of geometric arrangements in vector spaces of the embeddings of different languages. For translating a new source word, they map the corresponding word embedding to the target space using the learned mapping and find the nearest target word. In their approach, they found that simple linear mapping works better than non-linear mappings with multi-layer neural networks.  

\newcite{XingWLL15} enforce the word vectors to be of unit length during the learning of the embeddings and modify the objective function for learning the mapping to maximize the cosine similarity instead of using Euclidean distance.
To preserve length normalization after mapping, they enforce the \emph{orthogonality} constraint on the mapper. Instead of learning a mapping from the source to the target embedding space, \newcite{Faruqui14} use a technique based on Canonical Correlation Analysis (CCA) to project both source and target embeddings to a common low-dimensional space, where the correlation of the word pairs in the seed dictionary is maximized.
\newcite{artetxe2016emnlp} show that the above methods are variants of the same core optimization objective and propose a closed form solution for the mapper under orthogonality constraint. \newcite{SmithICLR17} find that this solution is closely related to the orthogonal \emph{Procrustes} solution. 
In their follow-up work, \newcite{artetxe2017acl} obtain competitive results using a seed dictionary of only 25 word pairs. They propose a \textit{self-learning framework} that performs two steps iteratively until convergence. In the first step, they use the dictionary (starting with the seed) to learn a linear mapping, which is then used in the second step to induce a new dictionary.

A more recent line of research attempts to eliminate the seed dictionary totally and learn the mapping in a purely unsupervised way. This was first proposed by \newcite{Valerio16}, who initially used an adversarial network similar to \newcite{conneau2018word}, and found that the mapper (which is also the encoder) translates everything to a single embedding, known commonly as the \emph{mode collapse} issue \cite{Goodfellow17}. To preserve diversity in mapping, he used a decoder to reconstruct the source embedding from the \emph{mapped embedding}, extending the framework to an adversarial autoencoder. His preliminary qualitative analysis shows encouraging results but not competitive with methods using bilingual seeds. He suspected issues with training and with the isomorphic assumption. In our work, we successfully address these issues with an improved model that also relaxes the isomorphic assumption. Our model uses two separate autoencoders, one for each language, which allows us to put more constraints to guide the mapping. We also distinguish the role of an encoder from the role of a mapper. The encoder projects embeddings to latent code vectors, which are then translated by the mapper.     

\newcite{Zhang17} improved adversarial training with orthogonal parameterization and cycle consistency. To aid training, they incorporate additional techniques like noise injection which works as a regularizer. For selecting the best model, they rely on sharp drops of the discriminator accuracy. In their follow-up work \cite{Zhang-17-emnlp}, they minimize Earth-Mover's distance between the distribution of the transformed source embeddings and the distribution of the target embeddings. \newcite{conneau2018word} show impressive results with adversarial training and refinement with the Procrustes solution. Instead of using the adversarial loss, \newcite{Xu2018} use Sinkhorn distance and adopt cycle consistency inspired by the CycleGAN \cite{CycleGAN2017}. We also incorporate cycle consistency along with the adversarial loss. However, while all these methods learn the mapping in the original embedding space, our approach learns it in the latent code space considering both the mapper and the target encoder as adversary. In addition, we use a post-cycle reconstruction to guide the mapping. 

A number of non-adversarial methods have also been proposed recently. \newcite{Artetxe-2018-acl} learn an initial dictionary by exploiting the structural similarity of the embeddings and use a robust self-learning algorithm to improve it iteratively. \newcite{Hoshen-18} align the second moment of word distributions of the two languages using principal component analysis (PCA) and then refine the alignment iteratively using a variation of the Iterative Closest Point (ICP) method used in computer vision. \newcite{david2018gromov} cast the problem as an optimal transport  problem and exploit the Gromov-Wasserstein distance which measures how similarities between pairs of words relate across languages.


 

\section{Approach}
\label{sec:model}
Let $\Xs = \{x_1, \ldots, x_n\}$ and $\Ys = \{y_1, \ldots, y_m\}$ be two sets consisting of $n$ and $m$ word embeddings of $d$-dimensions for a source and a target language, respectively. We assume that $\Xs$ and $\Ys$ are trained independently from monolingual corpora. Our aim is to learn a mapping $f(x)$ in an unsupervised way (\ie\ no bilingual dictionary given) such that for every $x_i$, $f(x)$ corresponds to its translation in $\Ys$. Our overall approach follows the same sequence of steps as \newcite{conneau2018word}: 
\begin{enumerate}
    \item Induction of seed dictionary through adversarial training. \vspace{-.8em}
    \item Iterative refinement of the initial mapping through the Procrustes solution. \vspace{-.8em}
    \item Apply CSLS for nearest neighbor search.
\end{enumerate}
\noindent We propose a novel adversarial autoencoder model to learn the initial mapping for inducing a seed dictionary in step (i), and we adopt existing refinement methods for steps (ii) and (iii).


\subsection{Adversarial Autoencoder for Initial Dictionary Induction} \label{subsec:adversary}

\begin{figure}[t!]
  \centering
\scalebox{0.98}{
  \includegraphics[width=1\linewidth,trim=4 4 4 4,clip]{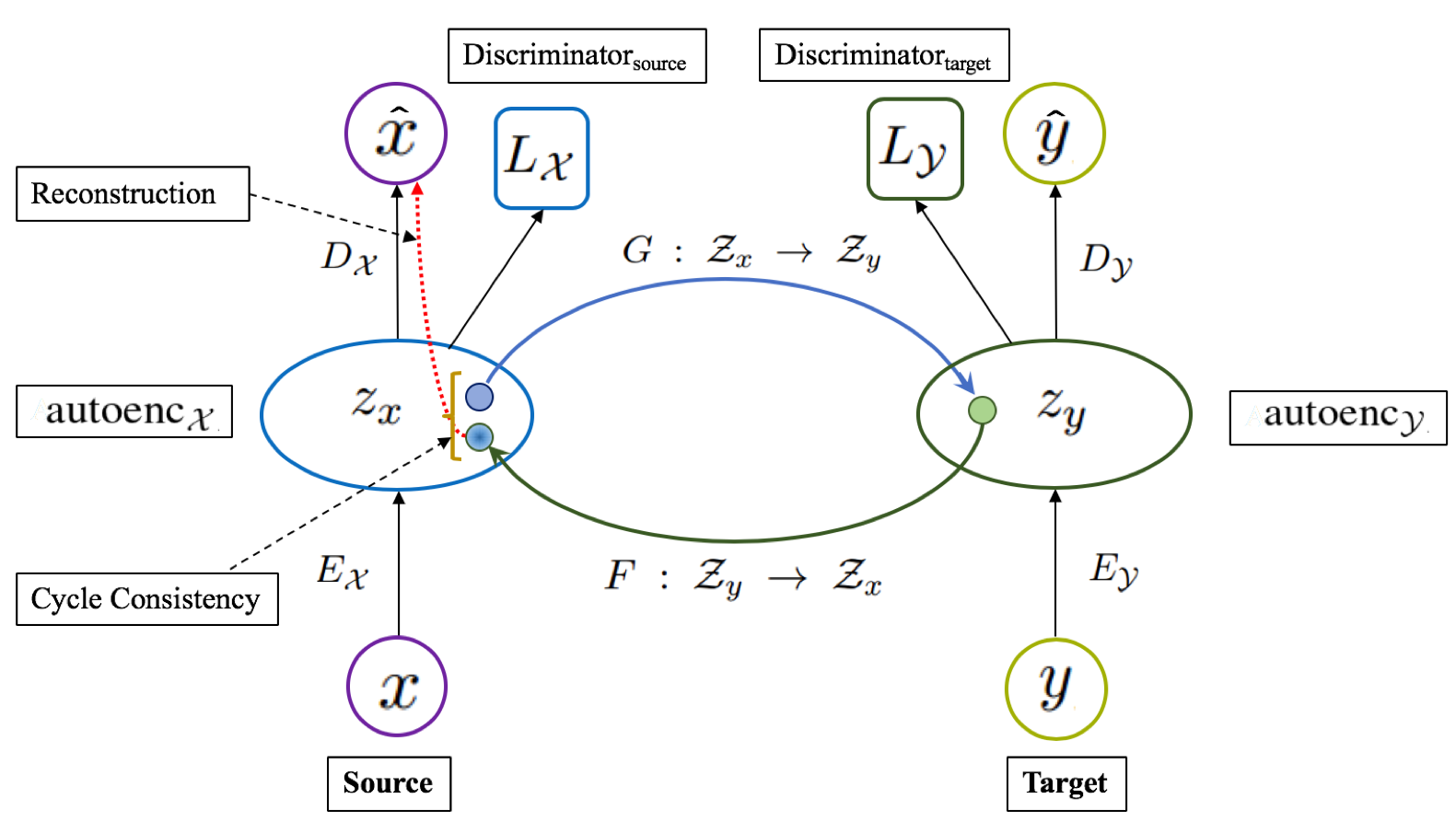}}
\caption{Our proposed adversarial autoencoder framework for unsupervised word translation.}
\label{fig:proposed-model}
\end{figure}

Our proposed model (Figure \ref{fig:proposed-model}) has two \textbf{autoencoders}, one for each language. Each autoencoder comprises an encoder $E_{\Xs}$ (\resp\ $E_{\Ys}$) and a decoder $D_{\Xs}$ (\resp\ $D_{\Ys}$). The encoders transform an input $x$ (\resp\ $y$) into a latent code $z_x$ (\resp\ $z_y$) from which the decoders try to reconstruct the original input. We use a linear encoder and $l_2$ \textbf{reconstruction loss}

\begin{align}
z_{x_i} = \theta_{E_\Xs} x_i; \hspace{1.5em} \hat{x}_i = \theta_{D_\Xs} z_{x_i}\\
\mathcal{L}_{\text{autoenc}_\Xs}(\theta_{E_\Xs},\theta_{D_\Xs}) = \frac{1}{n} \sum_{i=1}^{n}  \| {x_i} - \hat{x}_i \|^2 \label{autoenc1loss} 
\end{align}
\normalsize


\noindent where $\theta_{E_\Xs} \in \mathbb{R}^{c \times d}$ and $\theta_{D_\Xs} \in \mathbb{R}^{d \times c}$ are the parameters of the encoder and the decoder for $d$-dimensional word embedding and $c$-dimensional code vector.\footnote{We also experimented with a non-linear encoder, but it did not work well.}  The encoder, decoder and the reconstruction loss for the other autoencoder ($\text{autoenc}_\Ys$) is similarly defined. 

Let $q(z_x|x)$ and $q(z_y|y)$ be the encoding distributions of the two autoencoders. We use \emph{adversarial training} to find a mapping between $q(z_x|x)$ and $q(z_y|y)$. This is in contrast with most existing methods (\eg\ \newcite{conneau2018word,artetxe2017acl}) that directly map the distribution of the source word embeddings $p(x)$ to the distribution of the target $p(y)$. As \newcite{Anders-18} pointed out, the \emph{isomorphism} does not hold in general between the word embedding spaces of two languages. {Mapping the latent codes gives our model more flexibility to induce the required semantic structures in its code space that could potentially yield more accurate mappings.} 

As shown in Figure \ref{fig:proposed-model}, we include two linear \textbf{mappings} $G:\Zs_x\rightarrow\Zs_y$ and $F:\Zs_y\rightarrow\Zs_x$ to project the code vectors (samples from $q(.|.)$) from one language to the other. In addition, we have two language \textbf{discriminators}, $L_\Xs$ and $L_\Ys$. The discriminators are trained to discriminate between the mapped codes and the encoded codes, while the mappers and encoders are jointly trained to fool their respective discriminator. This results in a \textbf{three-player} game, where the discriminator tries to identify the origin of a code, and the mapper and the encoder act together to prevent the discriminator to succeed by making the mapped vector and the encoded vector as similar as possible. 


\paragraph{Discriminator Loss} Let $\theta_{L_\Xs}$ and $\theta_{L_\Ys}$ denote the parameters of the two discriminators, and $W_{G}$ and $W_{F}$ are the mapping weight matrices. The  loss for the  source discriminator $L_\Xs$ can be written as      

\small
\begin{align}
\hspace{-0.5em}\mathcal{L}_{L_\Xs}(\theta_{L_\Xs}|W_{F},\theta_{E_\Xs}) &= - \frac{1}{m} \sum_{j=1}^{m} \log P_{L_\Xs}(\text{src}=0|F(z_{y_j})) \nonumber \\
 & \hspace{0.8em} - \frac{1}{n} \sum_{i=1}^n \log P_{L_\Xs}(\text{src}=1|{z_{x_i}}) 
\label{discriminator1loss}
\end{align}
\normalsize

\noindent where $P_{L_\Xs}(\text{src}|z)$ is the probability according to $L_\Xs$ to distinguish whether $z$ is coming from the source encoder ($\text{src}=1$) or from the target-to-source mapper $F$ ($\text{src}=0$). The discrimination loss $\mathcal{L}_{L_\Ys}(\theta_{L_\Ys}|W_{G},\theta_{E_\Ys})$ is similarly defined for the target discriminator $L_\Ys$ using $G$ and ${E_\Ys}$.

Our discriminators have the same architecture as \newcite{conneau2018word}. It is a feed-forward network with two hidden layers of size 2048 and Leaky-ReLU activations. We apply dropout with a rate of 0.1 on the input to the discriminators. {Instead of using 1 and 0, we also apply a smoothing coefficient ($s=0.2$) in the discriminator loss.}   

\paragraph{Adversarial Loss} The mappers and encoders are trained jointly with the following adversarial loss to fool their respective discriminators.

\small
\begin{align}
\hspace{-0.5em}\mathcal{L}_{\text{adv}}(W_{F}, \theta_{E_\Xs}|\theta_{L_\Xs}) &= - \frac{1}{m} \sum_{i=1}^{m} \log P_{L_\Xs}(\text{src}=1|F(z_{y_j})) \nonumber \\
& \hspace{0.8em} -\frac{1}{n} \sum_{i=1}^n \log P_{L_\Xs}(\text{src}=0|z_{x_i}) 
\label{adversaryAloss}
\end{align}

\normalsize

\noindent The adversarial loss for mapper $G$ and encoder $E_\Ys$ is similarly defined. Note that we consider both the mapper and the target encoder as generators. This is in contrast to existing adversarial methods, which do not use any autoencoder in the target side. The mapper and the target encoder team up to fool the discriminator. {This forces the discriminator to improve its skill and vice versa for the generators, forcing them to produce indistinguishable codes through better mapping.}    




\paragraph{Cycle Consistency and Reconstruction}

The adversarial method introduced above maps a ``bag'' of source embeddings to a ``bag'' of target embeddings, and in theory, the mapper can match the target language distribution. However, mapping at the bag level is often insufficient to learn the individual word level mappings. In fact, there exist infinite number of possible mappings that can match the same target distribution. Thus to learn better mappings, we need to enforce more constraints to our objective. 

The first form of constraints we consider is \textbf{cycle consistency} to ensure that a source code $z_x$ translated to the target language code space, and {translated back} to the original space remains unchanged, \ie\  $z_x$$\rightarrow$$G(z_x)$$\rightarrow$$F(G(z_x))$$\approx$$z_x$. Formally, the cycle consistency loss in one direction:      

\small
\vspace{-1.0em}
\begin{align}
\mathcal{L}_{\text{cyc}}(W_{G}, W_{F})= \frac{1}{n} \sum_{i=1}^{n}  \| {z_{x_i}} - F(G(z_{x_i})) \| 
\label{cycleloss}
\end{align}
\normalsize

\noindent The loss in the other direction ($z_y \rightarrow F(z_y) \rightarrow G(F(z_y)) \approx z_y$) is similarly defined. 
In addition to cycle consistency, we include another constraint to guide the mapping further. In particular, we ask the decoder of the respective autoencoder to reconstruct the original input from the back-translated code. We compute this \textbf{post-cycle reconstruction loss} for the source autoencoder as follows:   

\small
\vspace{-1.0em}
\begin{align}
\mathcal{L}_{\text{rec}}(\theta_{E_\Xs}, \theta_{D_\Xs}, W_{G}, W_{F})  \hspace{-0.1em} = \frac{1}{n} \sum_{i=1}^{n}  \hspace{-0.1em} \| \hspace{-0.1em} {x_i} - \hspace{-0.2em} D_\Xs(F(G({z_{x_i}}))) \|^2 
\label{reconsAloss}
\end{align}
\normalsize

\noindent The reconstruction loss at the target autoencoder is defined similarly. Apart from improved mapping, both cycle consistency and reconstruction lead to more stable training in our experiments. Specifically, they help our training to converge and get around the \emph{mode collapse} issue \cite{Goodfellow17}. Since the model now has to translate the mapped code back to the source code and reconstruct the original word embedding, the generators cannot get away by mapping all source codes to a single target code. 


\paragraph{Total Loss} 

The total loss for mapping a batch from source to target is  

\vspace{-1.0em}
\begin{align}
\mathcal{L}_{\text{src} \rightarrow \text{tar}} = \mathcal{L}_{\text{adv}} + \lambda_1  \mathcal{L}_{\text{cyc}} + \lambda_2 \mathcal{L}_{\text{rec}} \label{eq:total}
\end{align}

\noindent where $\lambda_1$ and $\lambda_2$ control the relative importance of the three loss components. Similarly we define the total loss for mapping in the opposite direction $\mathcal{L}_{\text{tar} \rightarrow \text{src}}$. The complete objective of our model is:

\vspace{-1.0em}
\begin{align}
\mathcal{L}_{\text{total}} = \mathcal{L}_{\text{src} \rightarrow \text{tar}} + \mathcal{L}_{\text{tar} \rightarrow \text{src}} 
\end{align}

\subsection{Training and Dictionary Construction}

We present the training procedure of our model and the overall word translation process in Algorithm \ref{alg:training}. We first pre-train the autoencoders separately on monolingual embeddings (Step 1). This pre-training is required to induce word semantics (and relations) in the latent code space. 

We start adversarial training (Step 2) by updating the discriminators for $n\_critics$ (5) times, each time with a random batch. Then we update the generators (the mapper and target encoder) on the adversarial loss. The mappers then go through two more updates, one for cycle consistency and another for post-cycle reconstruction. The autoencoders (encoder-decoder) in this stage get updated only on the post-cycle reconstruction loss. We also apply the orthogonalization update to the mappers following \newcite{conneau2018word} with $\beta=0.01$.

\begin{algorithm2e}[h!]
\footnotesize
\SetKwInOut{Input}{Input}\SetKwInOut{Output}{Output}
\SetAlgoNoLine
\SetNlSkip{0em}
\Input{Two sets of word embeddings: $\Xs$ and $\Ys$}
\tcp{Initial autoencoder training}
1. Train \text{autoenc}$_\Xs$ and \text{autoenc}$_\Ys$ separately for some epochs on monolingual embeddings  (Eq. \ref{autoenc1loss}); \\
\tcp{Adversarial training}
2.  \For{n\_epochs}    
    { 
        \For{n\_iterations}
        {\tcp{Critic update}
         \For{n\_critics} 
        	{
            \hspace{-1em} \Ni Sample a batch from $\Xs$ and $\Ys$ \\
            \hspace{-1em} \Nii Update discriminators ($L_\Xs$, $L_\Ys$) (Eq. \ref{discriminator1loss})\\
            }
            \hspace{-1em} \Na Sample a batch from $\Xs$ as source and $\Ys$ as target \\ 
            \hspace{-1em} \Nb Update mapper $G$ and encoder $E_\Ys$ on adversarial loss to fool $L_\Ys$ (Eq. \ref{adversaryAloss})\\
            \hspace{-1em} \Nc Update mappers $G$ and $F$ on cycle consistency loss (Eq. \ref{cycleloss})\\
            \hspace{-1em} \Nd Update mappers ($G$, $F$) and $\text{autoenc}_\Xs$ on post-cycle reconstruction loss (Eq. \ref{reconsAloss}) \\
            \hspace{-1em}\tcp{Orthogonalize the mapper}
            \hspace{-1em} 
            \Ne Update weight matrices of mapper $G$ and $F$ using:  \hspace*{0.9cm} $W_{G} \leftarrow (1+\beta)W_{G} - \beta(W_{G}W_{G}^T)W_{G}$ \\
            \hspace*{1cm} $W_{F} \leftarrow (1+\beta)W_{F} - \beta(W_{F}W_{F}^T)W_{F}$\\
            \hspace{-1em} \Nf Sample a batch from $\Ys$ as source and $\Xs$ as target and update accordingly (symmetric to \Nb-\Ne steps). 
        }
        Use \textit{validation criterion} to save the best model.
    }
\tcp{Iterative Procrustes/fine-tuning}
3. Load the best model.\\
\For{n\_iterations}    
        { 
            \hspace{-1em} \Na Build a synthetic dictionary (using source encoder, source-to-target mapper, and CSLS)\\
            \hspace{-1em} \Nb Apply the Procrustes solution on the dictionary.
         }
\tcp{Test}
4. Test the model on gold bilingual dictionary.\\

\caption{Unsupervised word translation with cycle-consistent adversarial autoencoder}
\label{alg:training}
\end{algorithm2e}
\normalsize

Our training setting is similar to \newcite{conneau2018word}, and we apply the same pre- and post-processing steps. We use stochastic gradient descent (SGD) with a batch size of 32, a learning rate of 0.1, and a decay of 0.98.

For selecting the best model, we use the \textit{unsupervised validation criterion} proposed by \newcite{conneau2018word}, which correlates highly with the mapping quality. In this criterion, $10,000$ most frequent source words along with their nearest neighbors in the target space are considered. The average cosine similarity between these pseudo translations is considered as the validation metric. 

The initial bilingual dictionary induced by adversarial training (or any other unsupervised method) is generally of lower quality than what could be achieved by a supervised method. \newcite{conneau2018word} and \newcite{Artetxe-2018-acl} propose fine-tuning methods to refine the initial mappings. Similar to  \newcite{conneau2018word}), we fine-tune our initial mappings ($G$ and $F$) by iteratively solving the \emph{Procrustes} problem and applying a dictionary induction step.
This method uses singular value decomposition or SVD of $\Zs_y^T \Zs_x$ to find the optimal mappings $G$ (similarly SVD($\Zs_x^T {\Zs_y})$ for $F$) given the approximate alignment of words from the previous step. For generating synthetic dictionary in each iteration, we only consider the translation pairs that are mutual nearest neighbors. In our fine-tuning, we run five iterations of this process. For finding the nearest neighbors, we use the Cross-domain Similarity Local Scaling (CSLS)  which works better in mitigating the hubness problem \cite{conneau2018word}. 



\section{Experimental Settings}
\label{sec:setting}
Following the tradition, we evaluate our model on \textbf{word translation} (\aka\ \textbf{bilingual lexicon induction}) task, which measures the accuracy of the predicted dictionary to a gold standard dictionary.

\subsection{Datasets}

We evaluate our model on two different datasets. The first one is from \newcite{conneau2018word}, which consists of FastText monolingual embeddings of ($d$ =) 300 dimensions \cite{bojanowski2017enriching} trained on Wikipedia monolingual corpus and gold dictionaries for 110 language pairs.\footnote{\url{https://github.com/facebookresearch/MUSE}} To show the generality of different methods, we consider \textbf{European}, \textbf{non-European} and \textbf{low-resource} languages. In particular, we evaluate on English (En) from/to Spanish (Es), German (De), Italian (It), Arabic (Ar), Malay (Ms), and Hebrew (He).    



We also evaluate on the more challenging dataset of \newcite{Dinu-iclr-workshop15} and its subsequent extension by \newcite{artetxe2018aaai}. We will refer to this dataset as \textbf{Dinu-Artexe} dataset. From this dataset, we choose to experiment on English from/to Italian and Spanish. English and Italian embeddings were trained on WacKy corpora using CBOW \cite{Mikolov-word2vec}, while the Spanish embeddings were trained on WMT News Crawl. The CBOW vectors are also of 300 dimensions. 

\subsection{Baselines}

We compare our method with the \textbf{unsupervised} models of \newcite{conneau2018word},  \newcite{Artetxe-2018-acl},  \newcite{david2018gromov}, \newcite{Xu2018}, and \newcite{Hoshen-18}. 

To evaluate how our unsupervised method compares with methods that rely on a bilingual seed dictionary, we follow \newcite{conneau2018word}, and compute a \textbf{supervised} baseline that uses the Procrustes solution directly on the seed dictionary (5000 pairs) to learn the mapping function, and then uses CSLS to do the nearest neighbor search. We also compare with the supervised approaches of \newcite{artetxe2017acl,artetxe2018aaai}, which to our knowledge are the state-of-the-art supervised systems. 
For some of the baselines, results are reported from their papers, while for the rest we report results by running the publicly available codes on our machine.

For training our model on European languages, the weight for cycle consistency ($\lambda_1$) in Eq. \ref{eq:total} was always set to 5, and the weight for post-cycle reconstruction ($\lambda_2$) was set to 1. For non-European languages, we use different values of $\lambda_1$ and $\lambda_2$ for different language pairs.
\footnote{We did not tune the $\lambda$ values much, rather used our initial observation. Tuning $\lambda$ values might yield even better results.} The dimension of the code vectors in our model was set to 350.

\section{Results}
\label{sec:result}
We present our results on European languages on the datasets of \newcite{conneau2018word} and \newcite{Dinu-iclr-workshop15} in Tables \ref{tab:european-results} and \ref{tab:european-results-dinu}, while the results on non-European languages are shown in Table \ref{tab:non-europian}. Through experiments, our goal is to assess: 
\begin{enumerate}
    \item Does the unsupervised mapping method based on our proposed adversarial autoencoder model improve over the best existing adversarial method of \newcite{conneau2018word} in terms of mapping accuracy and convergence (Section \ref{subsec:with-conn})? 
    \item How does our unsupervised mapping method compare with other unsupervised and supervised approaches (Section \ref{subsec:exist})?
    \item Which components of our adversarial autoencoder model attribute to improvements (Section \ref{subsec:dissec})?
\end{enumerate}

\subsection{Comparison with \newcite{conneau2018word}} \label{subsec:with-conn}

Since our approach follows the same steps as \newcite{conneau2018word}, we first compare our proposed model with their model on European (Table \ref{tab:european-results}), non-European and low-resource languages (Table \ref{tab:non-europian}) on their dataset. In the tables, we present the numbers that they reported in their paper (\newcite{conneau2018word} (paper)) as well as the results that we get by running their code on our machine (\newcite{conneau2018word} (code)). For a fair comparison with respect to the quality of the learned mappings (or induced seed dictionary), here we only consider the results of our approach that use the refinement procedure of \newcite{conneau2018word}.

\begingroup
\setlength{\tabcolsep}{3pt}
\begin{table}[t!]
\centering
\footnotesize
\scalebox{0.85}{\begin{tabular}{l|cccccc}
\toprule
&\multicolumn{2}{c}{\textbf{En-Es}}  & \multicolumn{2}{c}{\textbf{En-De}} & \multicolumn{2}{c}{\textbf{En-It}} 
\\
& \textbf{$\rightarrow$} & \textbf{$\leftarrow$} & \textbf{$\rightarrow$} & \textbf{$\leftarrow$} & 
\textbf{$\rightarrow$} & 
\textbf{$\leftarrow$} 
\\       
\midrule
\textbf{Supervised} (Procrustes-CSLS)      & 82.4 & 83.9 & 75.3 & 72.7 & 78.1 & 78.1 \\
\midrule
\textbf{Unsupervised Baselines} & \\
\newcite{Artetxe-2018-acl}  & 82.2 & {84.4} & 74.9 & {74.1} & {78.8} & 79.5 \\
\newcite{alvarezmelis2018gromov}  & 81.7 & 80.4 &  71.9 & 72.8 & {78.9} & 75.2 \\
\newcite{Ruochen-emnlp18} 		  & 79.5 & 77.8 & 69.3 & 67.0 & 73.5 &  72.6\\
\newcite{Hoshen-18} 				  & 82.1 & 84.1 & 74.7 & 73.0 & 77.9 & 77.5 \\
\newcite{conneau2018word} (paper) & 81.7 & 83.3 & 74.0 & 72.2 & - & - \\
\newcite{conneau2018word} (code)  & 82.3 & 83.7 & 74.2 & 72.6 & 78.3 & 78.1\\
\midrule
\textbf{Our Unsupervised Approach} & \\
Adversarial autoencoder + & \\
\quad \newcite{conneau2018word} Refinement  &  82.6 & {84.4} & \textbf{75.5} & 73.9 & 78.8 & {78.5} \\
\quad \newcite{Artetxe-2018-acl} Refinement  &  \textbf{82.7} & \textbf{84.7} & {75.4} & \textbf{74.3} & \textbf{79.0} & \textbf{79.6} \\

\bottomrule
\end{tabular}}
\caption{\textbf{Word translation accuracy (P@1)} on \textbf{European} languages on the dataset of \newcite{conneau2018word} using \textbf{fastText} embeddings (trained on \textbf{Wikipedia}). `-' indicates the authors did not report the number. 
} 
\label{tab:european-results} 
\end{table}
\endgroup

In Table \ref{tab:european-results}, we see that our \textbf{Adversarial autoencoder + \newcite{conneau2018word} Refinement} outperforms \newcite{conneau2018word} in all the six translation tasks involving European language pairs, yielding gains in the range 0.3 - 1.3\%. Our method is also superior to theirs for the non-European and low-resource language pairs in Table \ref{tab:non-europian}. Here our method gives more gains ranging from 1.8 to 4.3\%. Note specifically that Malay (Ms) is a low-resource language, and the FastText contains word vectors for only 155K Malay words. We found their model to be very fragile for En from/to Ms, and does not converge at all for Ms$\rightarrow$En. We ran their code 10 times for Ms$\rightarrow$En but failed every time. Compared to that, our method is more robust and converged most of the time we ran.

\begingroup
\setlength{\tabcolsep}{3pt}
\begin{table}[t!]
\centering
\footnotesize
\scalebox{0.85}{\begin{tabular}{l|cccccc}
\toprule
&\multicolumn{2}{c}{\textbf{En-Ar}}  & \multicolumn{2}{c}{\textbf{En-Ms}} & \multicolumn{2}{c}{\textbf{En-He}} 
\\
& \textbf{$\rightarrow$} & \textbf{$\leftarrow$} & \textbf{$\rightarrow$} & \textbf{$\leftarrow$} & 
\textbf{$\rightarrow$} & 
\textbf{$\leftarrow$} 
\\       
\midrule
\textbf{Supervised Baselines} &  \\
\newcite{artetxe2017acl} & 24.8 & 43.3 & 38.8 & 41.6 & 32.7 & 51.1 \\
\newcite{artetxe2018aaai} & 36.2 & 52.9 & 51.2 & 47.7 & 43.6 & 56.8 \\
Supervised (Procrus-CSLS)  & 34.5 & 49.7 & 47.3 & 46.6 & 39.2 & 54.1
\\
\midrule
\textbf{Unsupervised Baselines} & \\
\newcite{Hoshen-18} & 34.4 & 49.3 & ** & ** & 36.5 & 52.3\\
\newcite{Artetxe-2018-acl} & 36.1 & {48.7} & 54.0 &  \textbf{55.4} & {43.8} &  \textbf{57.5} \\
\newcite{conneau2018word} (code) & 29.3 & 47.6 & 46.2 & ** & 36.8 & 53.1 \\
\midrule
\textbf{Our Unsupervised Approach} & \\ 
Adversarial autoencoder + & \\
\quad \newcite{conneau2018word} Refinement  & {33.6} & 49.7 & {49.5} & 44.3 & 40.0 & 54.9 \\
\quad \newcite{Artetxe-2018-acl} Refinement    &  \textbf{36.3} & \textbf{52.6} & \textbf{54.1} & {51.7} & \textbf{44.0} & {57.1} \\
\bottomrule
\end{tabular}}
\caption{\textbf{Word translation accuracy (P@1)} on \textbf{non-European} and \textbf{low-resource} languages on the dataset of \newcite{conneau2018word}  using \textbf{fastText} embeddings. ** indicates the model failed to converge.}
\label{tab:non-europian} 
\end{table}
\endgroup

If we compare our method with the method of \newcite{conneau2018word} on the more challenging Dinu-Artexe dataset in Table \ref{tab:european-results-dinu}, we see here also our method performs better than their method in all the four translation tasks involving European language pairs. In this dataset, our method shows more robustness compared to their method. For example, their method had difficulties in converging for En from/to Es translations; for En$\rightarrow$Es, it converges only 2 times out of 10 attempts, while for Es$\rightarrow$En it did not converge a single time in 10 attempts. Compared to that, our method was more robust, converging 4 times out of 10 attempts. 

In Section \ref{subsec:dissec}, we compare our model with \newcite{conneau2018word} more rigorously by evaluating them with and without fine-tuning and measuring their performance on P@1, P@5, and P@10.

\subsection{Comparison with Other Methods} \label{subsec:exist}

In this section, we compare our model with other state-of-the-art methods that do not follow the same procedure as us and \newcite{conneau2018word}. For example, \newcite{Artetxe-2018-acl} do the initial mapping in the similarity space, then they apply a different self-learning method to fine-tune the embeddings, and perform a final refinement with symmetric re-weighting. Instead of mapping from source to target, they map both source and target embeddings to a common space.

\begingroup
\setlength{\tabcolsep}{3pt}
\begin{table}[t!]
\centering
\footnotesize
\scalebox{0.95}{\begin{tabular}{l|cccc}
\toprule
&\multicolumn{2}{c}{\textbf{En-It}}  & \multicolumn{2}{c}{\textbf{En-Es}} 
\\
& \textbf{$\rightarrow$} & \textbf{$\leftarrow$} & \textbf{$\rightarrow$} & \textbf{$\leftarrow$} 
\\
\midrule
\textbf{Supervised Baselines} &  \\
\newcite{artetxe2017acl} & 39.7 & 33.8 & 32.4 & 27.2 \\
\newcite{artetxe2018aaai} & 45.3 & 38.5 & 37.2 & 29.6 \\
Procrustes-CSLS & 44.9 & 38.5 & 33.8 & 29.3\\
\midrule
\textbf{Unsupervised Baselines} & \\
\newcite{Artetxe-2018-acl}        & \textbf{47.9} & {42.3} & \textbf{37.5} & {31.2} \\
\newcite{conneau2018word} (paper) & 45.1 & 38.3 & - & - \\
\newcite{conneau2018word} (code)  & 44.9 & 38.7 & 34.7 & ** \\
\midrule
\textbf{Our Unsupervised Approach} & \\
Adversarial autoencoder + & \\
\quad \newcite{conneau2018word} Refinement          & 45.3 & 39.4 & 35.2 & 29.9   \\
\quad \newcite{Artetxe-2018-acl} Refinement    &  {47.6} & \textbf{42.5} & {37.4} & \textbf{31.9} \\
\bottomrule
\end{tabular}}
\caption{\textbf{Word translation accuracy (P@1)} on the \textbf{English-Italian} and \textbf{English-Spanish} language pairs  of \textbf{Dinu-Artetxe} dataset \cite{Dinu-iclr-workshop15,artetxe2017acl}. All methods use \textbf{CBOW} embeddings. ** indicates the model failed to converge; `-' indicates the authors did not report the number.}
\label{tab:european-results-dinu} 
\end{table}
\endgroup

Let us first consider the results for European language pairs on the dataset of \newcite{conneau2018word} in Table \ref{tab:european-results}. Our \textbf{Adversarial autoencoder + \newcite{conneau2018word} Refinement} performs better than most of the other methods on this dataset, achieving the highest accuracy for 4 out of 6 translation tasks. For De$\rightarrow$En, our result is very close to the best system of \newcite{Artetxe-2018-acl}  with only 0.2\% difference.

On the dataset of \newcite{Dinu-iclr-workshop15,artetxe2017acl} in Table \ref{tab:european-results-dinu}, our \textbf{Adversarial autoencoder + \newcite{conneau2018word} Refinement} performs better than other methods except \newcite{Artetxe-2018-acl}. On average our method lags behind by about 2\%. However, as mentioned, they follow a different refinement and mapping methods. 
For non-European and low-resource language pairs in Table \ref{tab:non-europian}, our \textbf{Adversarial autoencoder + \newcite{conneau2018word} Refinement} exhibits better performance than others in one translation task, where the model of \newcite{Artetxe-2018-acl} performs better in the rest. One important thing to notice here is that other unsupervised models (apart from ours and \newcite{Artetxe-2018-acl}) fail to converge in one or more language pairs.

We notice that the method of \newcite{Artetxe-2018-acl} gives better results than other baselines, even in some translation tasks they achieve the highest accuracy. To understand whether the improvements of their method are due to a better initial mapping or better post-processing, we conducted two additional experiments. In our first experiment, we use their method to induce the initial seed dictionary and then apply iterative Procrustes solution (same refinement procedure of \newcite{conneau2018word}) for refinement. Table \ref{tab:experiment-artetxe} shows the results. Surprisingly, on both datasets their initial mappings fail to produce any reasonable results. So we suspect that the main gain in \cite{Artetxe-2018-acl} comes from their fine-tuning method, which they call \textit{robust self learning}. In our second experiment, we use the initial dictionary induced by our adversarial training and then apply their refinement procedure. Here for most of the translation tasks, we achieve better results; see the model \textbf{Adversarial autoencoder + \newcite{Artetxe-2018-acl} Refinement} in Tables \ref{tab:european-results} - \ref{tab:european-results-dinu}. These two experiments demonstrate that the quality of the initial dictionary induced by our model is far better than that of \newcite{Artetxe-2018-acl}.


\begingroup
\setlength{\tabcolsep}{5pt}
\begin{table}[t!]
\centering
\footnotesize
\scalebox{1.00}{\begin{tabular}{l|cccc}
\toprule
&\multicolumn{2}{c}{\textbf{En-It}}  & \multicolumn{2}{c}{\textbf{En-Es}} 
\\
& \textbf{$\rightarrow$} & \textbf{$\leftarrow$} & \textbf{$\rightarrow$} & \textbf{$\leftarrow$} 
\\
\midrule
{Dinu-Artetxe} Dataset  & ** & ** & ** & ** \\
Conneau Dataset & 01.2 & 01.6 & 04.7 & 05.1 \\
\bottomrule
\end{tabular}}
\caption{\newcite{conneau2018word} refinement applied to the initial mappings of \newcite{Artetxe-2018-acl}. ** indicates the model failed to converge.}
\label{tab:experiment-artetxe} 
\end{table}
\endgroup



\begingroup
\setlength{\tabcolsep}{3pt}
\begin{table*}[t!]
\centering
\scalebox{0.76}{\begin{tabular}{l|ccc|ccc|ccc|ccc|ccc|ccc}
&\multicolumn{3}{c}{\textbf{En$\rightarrow$Es}} &\multicolumn{3}{c}{\textbf{Es$\rightarrow$En}} & \multicolumn{3}{c}{\textbf{En$\rightarrow$De}} & \multicolumn{3}{c}{\textbf{De$\rightarrow$En}} & \multicolumn{3}{c}{\textbf{En$\rightarrow$It}} & \multicolumn{3}{c}{\textbf{It$\rightarrow$En}}  \\
&\textbf{P@1} & \textbf{P@5} & \textbf{P@10} &\textbf{P@1} & \textbf{P@5} & \textbf{P@10} &\textbf{P@1} & \textbf{P@5} & \textbf{P@10} &\textbf{P@1} & \textbf{P@5} & \textbf{P@10} &\textbf{P@1} & \textbf{P@5} & \textbf{P@10} &\textbf{P@1} & \textbf{P@5} & \textbf{P@10} \\
\midrule
& \multicolumn{18}{c}{\textbf{Without Fine-Tuning}} \\[0.5em] 
\textbf{Conneau-18} & 65.3 & 73.8 & 80.6 & 66.7 & 78.3 & 80.8 & 61.5 & 70.1 & 78.2 & 60.3 & 70.2 & 77.0 & 64.8 & 75.3 & 79.4 & 63.8 & 77.1 & 81.8  \\
\midrule
\textbf{Our (full)}  & 71.8 & 81.1 & 85.7 & 72.7 & 81.5 & 83.8 & 64.9 & 74.4 & 81.8 & 63.1 & 71.3 & 79.8 & 68.2 & 78.9 & 83.7 & 67.5 & 77.6 & 82.1 
\\
~~- Enc. adv  & 70.5 & 79.7 & 83.5 & 71.3 & 80.4 & 83.3 & 63.7 & 73.5 & 79.3 & 62.6 & 70.5 & 79.0 & 67.6 & 77.3 & 82.7 & 66.2 & 78.3 & 82.5
\\
~~- -  {Recon}  & 70.1 &78.9 & 83.4 & 70.8	& 81.1 & 83.4 & 63.1 & 73.8 & 80.5 &  62.2 & 71.7 & 78.7 & 66.9 & 79.7 & 82.1 & 64.8 & 78.6 & 82.1
\\
~~- - - Cycle  &  66.8 & 76.5 & 82.1 & 67.2 & 79.9 & 82.7 & 61.4 & 69.7 & 77.8 & 60.1	 & 69.8 & 76.5 & 65.3 & 75.1 & 78.9 & 64.4 & 77.6 & 81.7
\\
\midrule
& \multicolumn{18}{c}{\textbf{With Fine-Tuning}} \\[0.5em] 
\textbf{Conneau-18} & 82.3 & 90.8 & 93.2 & 83.7 & 91.9 & 93.5 & 74.2 & 89.0 & 91.5 & 72.6 & 85.7 & 88.8 & 78.3 & 88.4 & 91.1 & 78.1 & 88.2 & 90.6
\\
\midrule
\textbf{Our (full)}  & 82.6 & 91.8 & 93.5 & 84.4 & 92.3 & 94.3 & 75.5 & 90.1 & 92.9 & 73.9 & 86.5 & 89.3 & 78.8 & 89.2 & 91.9 & 78.5 & 88.9 & 91.1\\
~~- Enc. adv  & 82.5 & 91.6 & 93.5 & 84.3 & 92.1 & 94.3 & 75.4 & 89.7 & 92.7 & 73.5 & 86.3 & 89.2 & 78.4 & 89.0 & 91.8 & 78.1 & 88.7 & 91.0
\\
~~- - {Recon}  & 82.5	& 91.6 & 93.4 & 84.1 & 92.2 & 94.3 & 75.3 & 89.4 & 92.6 & 73.2 & 85.9 & 89.0 & 78.2	& 89.1 & 91.9 & 78.2 & 88.8 & 91.2 
\\
~~- - - Cycle  & 82.4	& 91.0 & 93.1 & 83.6 & 92.2 & 94.0 & 74.3 & 89.7 & 92.6 & 72.7 & 86.1 & 89.1 & 77.8 & 89.2 & 91.8 & 77.4 & 88.3 & 90.8
\\
\bottomrule
\end{tabular}}
\caption{Ablation study of our adversarial autoencoder model on the dataset of \newcite{conneau2018word}.} 
\label{tab:ablation} 
\end{table*}
\endgroup

\subsection{Model Dissection} \label{subsec:dissec}
We further analyze our model by dissecting it and measuring the contribution of each novel component that is proposed in this work. We achieve this by \emph{incrementally} removing a new component from the model and evaluating it on different translation tasks. In order to better understand the contribution of each component, we evaluate each model by measuring its \textbf{P@1}, \textbf{P@5}, and \textbf{P@10} \textbf{with fine-tuning} and \textbf{without fine-tuning}. {In case of \textbf{without fine-tuning}, the models apply the CSLS neighbor search directly on the mappings learned from the adversarial training, \ie\ no Procrustes solution based refinement is done after the adversarial training.} This setup allows us to compare our model directly with the adversarial model of \newcite{conneau2018word}, putting the effect of fine-tuning aside.    

Table \ref{tab:ablation} presents the ablation results for En-Es, En-De, and En-It in both directions. The first row (\textbf{Conneau-18}) presents the results of \newcite{conneau2018word} that uses adversarial training to map the \emph{word embeddings}. The next row shows the results of \textbf{our full} model. The subsequent rows incrementally detach one component from our model. For example, \textbf{- Enc. adv} denotes the variant of our model where the target encoder is not trained on the adversarial loss ($\theta_{E_\Xs}$ in Eq. \ref{adversaryAloss}); \textbf{- - Recon} excludes the post-cycle reconstruction loss from \textbf{- Enc. adv}, and  \textbf{- - - Cycle} excludes the cycle consistency from \textbf{- - Recon}. Thus, \textbf{- - - Cycle} is a variant of our model that uses only adversarial loss to learn the mapping. However, it is important to note that in contrast to \newcite{conneau2018word}, our mapping is performed at the code space.          

As we compare our full model with the model of \newcite{conneau2018word} in the \emph{without fine-tuning} setting, we notice large improvements in all measures across all datasets: 5.1 - 7.3\% in En$\rightarrow$Es, 3 - 6\% in Es$\rightarrow$En, 3.4 - 4.3\% in En$\rightarrow$De, 1 - 3\% in De$\rightarrow$En,  3.4 - 4.3\% in En$\rightarrow$It, and 0.3 - 3.7\% in It$\rightarrow$En. These improvements demonstrate that our model finds a better mapping compared to \newcite{conneau2018word}. Among the three components, the cycle consistency is the most influential one across all languages. Training the target encoder adversarially also gives a significant boost. The reconstruction has less impact. If we compare the results of \emph{- - - Cycle} with \emph{Conneau-18}, we see sizeable gains for En-Es in both directions. This shows the benefits of mapping at the code level.    

Now let us turn our attention to the results with fine-tuning. Here also we see gains across all datasets for our model, although the gains are not as verbose as before (about 1\% on average). However, this is not surprising as it has been shown that \textit{iterative fine-tuning} with Procrustes solution is a robust method that can recover many errors made in the initial mapping \cite{conneau2018word}. Given a good enough initial mapping, the measures converge nearly to the same point even though the differences were comparatively more substantial initially; for example, notice that the scores are very similar for P@5 and P@10 measures after fine-tuning.


\section{Conclusions}
\label{sec:conclusion}
We have proposed an adversarial autoencoder framework to learn the cross-lingual mapping of monolingual word embeddings of two languages in a completely unsupervised way. In contrast to the existing methods that directly map word embeddings, our method first learns to transform the embeddings into latent code vectors by pretraining an autoencoder. 
We apply adversarial training to map the distributions of the source and target code vectors. In our adversarial training, both the mapper and the target encoder are treated as generators that act jointly to fool the discriminator. To guide the mapping further, we include constraints for cycle consistency and post-cycle reconstruction.

Through extensive experimentations on six different language pairs comprising European, non-European and low-resource languages from two different data sources, we demonstrate that our method outperforms the method of \newcite{conneau2018word} for all translation tasks in all measures (P@\{1,5,10\}) across all settings (with and without fine-tuning). Comparison with other existing methods also shows that our method learns better mapping (not considering the fine-tuning). With an ablation study, we further demonstrated that the cycle consistency is the most important component followed by the adversarial training of target encoder and the post-cycle reconstruction.  
In future work, we plan to incorporate knowledge from the similarity space in our adversarial framework.

\section*{Acknowledgments}

The authors would like to thank the funding support from MOE Tier-1 (Grant M4011897.020).

\bibliographystyle{acl_natbib}

\begin{thebibliography}{28}
\expandafter\ifx\csname natexlab\endcsname\relax\def\natexlab#1{#1}\fi

\bibitem[{Alvarez and Jaakkola(2018)}]{alvarezmelis2018gromov}
David Alvarez and Tommi Jaakkola. 2018.
\newblock \href {http://aclweb.org/anthology/D18-1214} {Gromov-wasserstein
  alignment of word embedding spaces}.
\newblock In \emph{Proceedings of the 2018 Conference on Empirical Methods in
  Natural Language Processing}, pages 1881--1890. Association for Computational
  Linguistics.

\bibitem[{Alvarez-Melis and Jaakkola(2018)}]{david2018gromov}
David Alvarez-Melis and Tommi Jaakkola. 2018.
\newblock \href {http://aclweb.org/anthology/D18-1214} {Gromov-wasserstein
  alignment of word embedding spaces}.
\newblock In \emph{Proceedings of the 2018 Conference on Empirical Methods in
  Natural Language Processing}, pages 1881--1890. Association for Computational
  Linguistics.

\bibitem[{Artetxe et~al.(2016)Artetxe, Labaka, and Agirre}]{artetxe2016emnlp}
Mikel Artetxe, Gorka Labaka, and Eneko Agirre. 2016.
\newblock \href {https://aclweb.org/anthology/D16-1250} {Learning principled
  bilingual mappings of word embeddings while preserving monolingual
  invariance}.
\newblock In \emph{Proceedings of the 2016 Conference on Empirical Methods in
  Natural Language Processing}, pages 2289--2294, Austin, Texas. Association
  for Computational Linguistics.

\bibitem[{Artetxe et~al.(2017)Artetxe, Labaka, and Agirre}]{artetxe2017acl}
Mikel Artetxe, Gorka Labaka, and Eneko Agirre. 2017.
\newblock \href {http://aclweb.org/anthology/P17-1042} {Learning bilingual word
  embeddings with (almost) no bilingual data}.
\newblock In \emph{Proceedings of the 55th Annual Meeting of the Association
  for Computational Linguistics (Volume 1: Long Papers)}, pages 451--462,
  Vancouver, Canada. Association for Computational Linguistics.

\bibitem[{Artetxe et~al.(2018{\natexlab{a}})Artetxe, Labaka, and
  Agirre}]{artetxe2018aaai}
Mikel Artetxe, Gorka Labaka, and Eneko Agirre. 2018{\natexlab{a}}.
\newblock \href
  {https://www.aaai.org/ocs/index.php/AAAI/AAAI18/paper/view/16935}
  {Generalizing and improving bilingual word embedding mappings with a
  multi-step framework of linear transformations}.
\newblock In \emph{Proceedings of the Thirty-Second AAAI Conference on
  Artificial Intelligence}, pages 5012--5019.

\bibitem[{Artetxe et~al.(2018{\natexlab{b}})Artetxe, Labaka, and
  Agirre}]{Artetxe-2018-acl}
Mikel Artetxe, Gorka Labaka, and Eneko Agirre. 2018{\natexlab{b}}.
\newblock \href {http://www.aclweb.org/anthology/P18-1073} {A robust
  self-learning method for fully unsupervised cross-lingual mappings of word
  embeddings}.
\newblock In \emph{ACL}.

\bibitem[{Bojanowski et~al.(2017)Bojanowski, Grave, Joulin, and
  Mikolov}]{bojanowski2017enriching}
Piotr Bojanowski, Edouard Grave, Armand Joulin, and Tomas Mikolov. 2017.
\newblock \href {http://aclweb.org/anthology/Q17-1010} {Enriching word vectors
  with subword information}.
\newblock \emph{Transactions of the Association for Computational Linguistics},
  5:135--146.

\bibitem[{Conneau et~al.(2018)Conneau, Lample, Ranzato, Denoyer, and
  J{\'e}gou}]{conneau2018word}
Alexis Conneau, Guillaume Lample, Marc'Aurelio Ranzato, Ludovic Denoyer, and
  Herv{\'e} J{\'e}gou. 2018.
\newblock \href {https://arxiv.org/pdf/1710.04087} {Word translation without
  parallel data}.
\newblock In \emph{International Conference on Learning Representations
  (ICLR)}.

\bibitem[{Dinu et~al.(2015)Dinu, Lazaridou, and Baroni}]{Dinu-iclr-workshop15}
Georgiana Dinu, Angeliki Lazaridou, and Marco Baroni. 2015.
\newblock \href {https://arxiv.org/abs/1412.6568} {Improving zero-shot learning
  by mitigating the hubness problem}.
\newblock In \emph{ICLR, Workshop track}.

\bibitem[{Faruqui and Dyer(2014)}]{Faruqui14}
Manaal Faruqui and Chris Dyer. 2014.
\newblock \href {https://doi.org/10.3115/v1/E14-1049} {Improving vector space
  word representations using multilingual correlation}.
\newblock In \emph{Proceedings of the 14th Conference of the European Chapter
  of the Association for Computational Linguistics}, pages 462--471.
  Association for Computational Linguistics.

\bibitem[{Goodfellow(2017)}]{Goodfellow17}
Ian~J. Goodfellow. 2017.
\newblock \href {http://arxiv.org/abs/1701.00160} {{NIPS} 2016 tutorial:
  Generative adversarial networks}.
\newblock \emph{CoRR}, abs/1701.00160.

\bibitem[{Hoshen and Wolf(2018)}]{Hoshen-18}
Yedid Hoshen and Lior Wolf. 2018.
\newblock \href {http://aclweb.org/anthology/D18-1043} {Non-adversarial
  unsupervised word translation}.
\newblock In \emph{Proceedings of the 2018 Conference on Empirical Methods in
  Natural Language Processing}, pages 469--478. Association for Computational
  Linguistics.

\bibitem[{Lample et~al.(2018{\natexlab{a}})Lample, Conneau, Denoyer, and
  Ranzato}]{lample2017unsupervised}
Guillaume Lample, Alexis Conneau, Ludovic Denoyer, and Marc'Aurelio Ranzato.
  2018{\natexlab{a}}.
\newblock \href {https://arxiv.org/abs/1711.00043} {Unsupervised machine
  translation using monolingual corpora only}.
\newblock In \emph{International Conference on Learning Representations
  (ICLR)}.

\bibitem[{Lample et~al.(2018{\natexlab{b}})Lample, Ott, Conneau, Denoyer, and
  Ranzato}]{lample2018phrase}
Guillaume Lample, Myle Ott, Alexis Conneau, Ludovic Denoyer, and Marc'Aurelio
  Ranzato. 2018{\natexlab{b}}.
\newblock \href {https://arxiv.org/abs/1804.07755} {Phrase-based \& neural
  unsupervised machine translation}.
\newblock In \emph{Proceedings of the 2018 Conference on Empirical Methods in
  Natural Language Processing (EMNLP)}.

\bibitem[{Luong et~al.(2015)Luong, Pham, and Manning}]{Luong15-bivec}
Thang Luong, Hieu Pham, and Christopher~D. Manning. 2015.
\newblock \href {https://doi.org/10.3115/v1/W15-1521} {Bilingual word
  representations with monolingual quality in mind}.
\newblock In \emph{Proceedings of the 1st Workshop on Vector Space Modeling for
  Natural Language Processing}, pages 151--159. Association for Computational
  Linguistics.

\bibitem[{Makhzani et~al.(2015)Makhzani, Shlens, Jaitly, and
  Goodfellow}]{Makhzani2015AdversarialA}
Alireza Makhzani, Jonathon Shlens, Navdeep Jaitly, and Ian~J. Goodfellow. 2015.
\newblock \href {https://arxiv.org/abs/1511.05644} {Adversarial autoencoders}.
\newblock \emph{CoRR}, abs/1511.05644.

\bibitem[{Miceli~Barone(2016)}]{Valerio16}
Antonio~Valerio Miceli~Barone. 2016.
\newblock \href {https://doi.org/10.18653/v1/W16-1614} {Towards cross-lingual
  distributed representations without parallel text trained with adversarial
  autoencoders}.
\newblock In \emph{Proceedings of the 1st Workshop on Representation Learning
  for NLP}, pages 121--126. Association for Computational Linguistics.

\bibitem[{Mikolov et~al.(2013{\natexlab{a}})Mikolov, Le, and
  Sutskever}]{Mikolov13}
Tomas Mikolov, Quoc~V. Le, and Ilya Sutskever. 2013{\natexlab{a}}.
\newblock \href {http://arxiv.org/abs/1309.4168} {Exploiting similarities among
  languages for machine translation}.
\newblock \emph{CoRR}, abs/1309.4168.

\bibitem[{Mikolov et~al.(2013{\natexlab{b}})Mikolov, Sutskever, Chen, Corrado,
  and Dean}]{Mikolov-word2vec}
Tomas Mikolov, Ilya Sutskever, Kai Chen, Greg~S Corrado, and Jeff Dean.
  2013{\natexlab{b}}.
\newblock \href
  {http://papers.nips.cc/paper/5021-distributed-representations-of-words-and-phrases-and-their-compositionality.pdf}
  {Distributed representations of words and phrases and their
  compositionality}.
\newblock In \emph{Advances in Neural Information Processing Systems 26}, pages
  3111--3119. Curran Associates, Inc.

\bibitem[{Ruder et~al.(2017)Ruder, Vulic, and Sogaard}]{ruder2017survey}
Sebastian Ruder, Ivan Vulic, and Anders Sogaard. 2017.
\newblock \href {http://arxiv.org/abs/1706.04902} {A survey of cross-lingual
  word embedding models}.

\bibitem[{Smith et~al.(2017)Smith, Turban, Hamblin, and Hammerla}]{SmithICLR17}
Samuel~L. Smith, David H.~P. Turban, Steven Hamblin, and Nils~Y. Hammerla.
  2017.
\newblock \href {http://arxiv.org/abs/1702.03859} {Offline bilingual word
  vectors, orthogonal transformations and the inverted softmax}.
\newblock In \emph{International Conference on Learning Representations
  (ICLR)}.

\bibitem[{S{\o}gaard et~al.(2018)S{\o}gaard, Ruder, and
  Vuli{\'{c}}}]{Anders-18}
Anders S{\o}gaard, Sebastian Ruder, and Ivan Vuli{\'{c}}. 2018.
\newblock \href {http://aclweb.org/anthology/P18-1072} {On the limitations of
  unsupervised bilingual dictionary induction}.
\newblock In \emph{Proceedings of the 56th Annual Meeting of the Association
  for Computational Linguistics (Volume 1: Long Papers)}, pages 778--788.
  Association for Computational Linguistics.

\bibitem[{Xing et~al.(2015)Xing, Wang, Liu, and Lin}]{XingWLL15}
Chao Xing, Dong Wang, Chao Liu, and Yiye Lin. 2015.
\newblock \href
  {https://pdfs.semanticscholar.org/77e5/76c02792d7df5b102bb81d49df4b5382e1cc.pdf}
  {Normalized word embedding and orthogonal transform for bilingual word
  translation.}
\newblock In \emph{HLT-NAACL}, pages 1006--1011. The Association for
  Computational Linguistics.

\bibitem[{Xu et~al.(2018{\natexlab{a}})Xu, Yang, Otani, and Wu}]{Xu2018}
Ruochen Xu, Yiming Yang, Naoki Otani, and Yuexin Wu. 2018{\natexlab{a}}.
\newblock \href {http://aclweb.org/anthology/D18-1268} {Unsupervised
  cross-lingual transfer of word embedding spaces}.
\newblock In \emph{Proceedings of the 2018 Conference on Empirical Methods in
  Natural Language Processing}, pages 2465--2474. Association for Computational
  Linguistics.

\bibitem[{Xu et~al.(2018{\natexlab{b}})Xu, Yang, Otani, and
  Wu}]{Ruochen-emnlp18}
Ruochen Xu, Yiming Yang, Naoki Otani, and Yuexin Wu. 2018{\natexlab{b}}.
\newblock \href {http://aclweb.org/anthology/D18-1268} {Unsupervised
  cross-lingual transfer of word embedding spaces}.
\newblock In \emph{Proceedings of the 2018 Conference on Empirical Methods in
  Natural Language Processing}, pages 2465--2474. Association for Computational
  Linguistics.

\bibitem[{Zhang et~al.(2017{\natexlab{a}})Zhang, Liu, Luan, and Sun}]{Zhang17}
Meng Zhang, Yang Liu, Huanbo Luan, and Maosong Sun. 2017{\natexlab{a}}.
\newblock \href {https://doi.org/10.18653/v1/P17-1179} {Adversarial training
  for unsupervised bilingual lexicon induction}.
\newblock In \emph{Proceedings of the 55th Annual Meeting of the Association
  for Computational Linguistics (Volume 1: Long Papers)}, pages 1959--1970.
  Association for Computational Linguistics.

\bibitem[{Zhang et~al.(2017{\natexlab{b}})Zhang, Liu, Luan, and
  Sun}]{Zhang-17-emnlp}
Meng Zhang, Yang Liu, Huanbo Luan, and Maosong Sun. 2017{\natexlab{b}}.
\newblock \href {https://doi.org/10.18653/v1/D17-1207} {Earth mover's distance
  minimization for unsupervised bilingual lexicon induction}.
\newblock In \emph{Proceedings of the 2017 Conference on Empirical Methods in
  Natural Language Processing}, pages 1934--1945. Association for Computational
  Linguistics.

\bibitem[{Zhu et~al.(2017)Zhu, Park, Isola, and Efros}]{CycleGAN2017}
Jun-Yan Zhu, Taesung Park, Phillip Isola, and Alexei~A Efros. 2017.
\newblock \href {https://arxiv.org/abs/1703.10593} {Unpaired image-to-image
  translation using cycle-consistent adversarial networks}.
\newblock In \emph{Computer Vision (ICCV), 2017 IEEE International Conference
  on}.

\end{thebibliography}

\end{document}